\documentclass{article}

\usepackage{graphicx}

\newtheorem{theorem}{Theorem}

\newtheorem{definition}[theorem]{Definition}

\input{tcilatex}
\begin{document}

\title{Syntactic sensitive complexity for symbol-free sequence}
\author{Cheng-Yuan Liou, Bo-Shiang Huang, Daw-Ran Liou, and Alex A. Simak \\
Department of Computer Science and Information Engineering, \\
National Taiwan University, Taiwan, Republic of China}
\date{}
\maketitle

\begin{abstract}
This work uses the L-system to construct a tree structure for the text
sequence and derives its complexity \cite{Liou10}. It serves as a measure of
structural complexity of the text. It is applied to anomaly detection in
data transmission.

\end{abstract}

\section{Introduction}

Complexity of the text has been developed with varying degrees of success, 
\cite{Koslicki11}\cite{Tino99}. This work devised a novel measure based on
L-system \cite{Liou10} that can compute the structural complexity of a text
sequence. Given a text, we first transform it into a binary string. Then,
use L-system to model the tree structure of this string and get its
structural complexity. We will introduce how to use L-system to model the
string in this section. The measure of complexity for the text sequence is
included in the next section. 

\subsection{Transforming binary string into rewriting rules}

The Lindenmayer system, or L-system, is a parallel rewriting system which
was introduced by the biologist Aristid Lindenmayer in 1968. The major
operation of L-system is rewriting. A set of rewriting rules, or
productions, are operated to define a complex object by successively
replacing parts of a simple initial object. The operations for a
hierarchical tree can be represented by a set of rewriting rules. These
rules can be further transformed into a bracketed string. A binary tree can
be represented by a bracketed string and the tree can be restored from the
string. This bracked string contains five symbols, $F$, $+$, $-$, $[$ , and $%
]$ . These symbols are defined in the following paragraph.

\begin{itemize}
\item $F$ denotes the current location of a tree node. It can be replaced by
any word or be omitted.

\item $+$ \ denotes the following string that represents the right subtree.

\item $-$ \ denotes the following string that represents the left subtree.

\item $[$ \ is pairing with $]$. \textquotedblleft $\lbrack \ldots ]$%
\textquotedblright\ denotes a subtree where \textquotedblleft $\dots $%
\textquotedblright\ indicates all the bracketed string of its subtree.
\end{itemize}

Given a binary tree, the direct way to represent it with L-system is to
construct rewriting rules that replace a tree with two smaller subtrees. An
example is shown in Fig.~\ref{fig:rewriting bracketed stringatcg without
nulls}. The left tree in Fig.~\ref{fig:rewriting bracketed stringatcg
without nulls} is expressed as $P\rightarrow \lbrack -F{T_{L}}][+F{T_{R}}]$.
The right tree is expressed as $P\rightarrow \lbrack -F{T_{L}}][+F{T_{R}}]$
and ${T_{R}}\rightarrow \lbrack -F{T_{R_{L}}}][+F{T_{R_{R}}}]$. We see that
the two rules, $P\rightarrow \lbrack -F{T_{L}}][+F{T_{R}}]$ and ${T_{R}}%
\rightarrow \lbrack -F{T_{R_{L}}}][+F{T_{R_{R}}}]$, are similar. The binary
tree example in Fig.\ref{fig:binary tree} can be transformed into the set of
rewriting rules in Table~\ref{tab:rewriting rules}.

\begin{figure}[tbp]
\centering
\includegraphics[width=4.7522in,height=1.7277in]{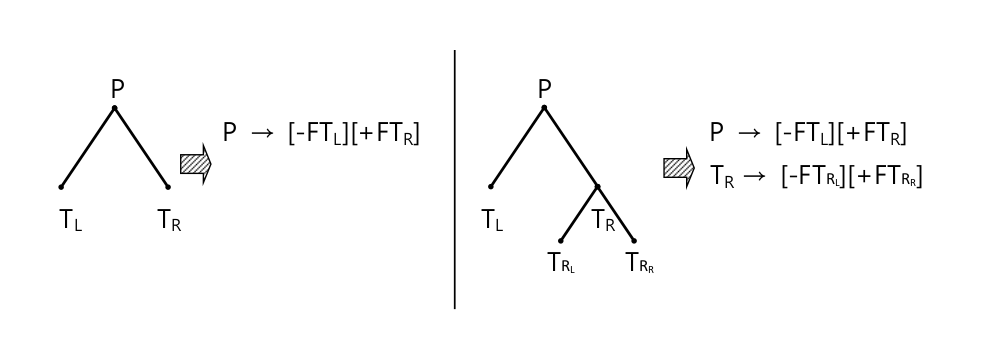}
\caption{Rewriting rules for the two
bracketed strings.}
\label{fig:rewriting bracketed stringatcg without nulls}
\end{figure}

Fig.\ref{fig:binary tree} shows an example with four fixed tree elements
representing the four 2-bit strings, \{ $00$, $10$, $01$, and $11$\}. In
this figure, every two leaves are combined into a small tree, and two small
trees are combined into a bigger tree recursively. In this way, we can get a
whole binary tree for a long binary string. Each node of the tree represents
all its descendant string sections. The binary tree example in Fig.\ref%
{fig:binary tree} can be transformed into the set of rewriting rules in
Table~\ref{tab:rewriting rules}.

\begin{figure}[tbp]
\centering
\includegraphics[width=2.1071in,height=1.3491in]{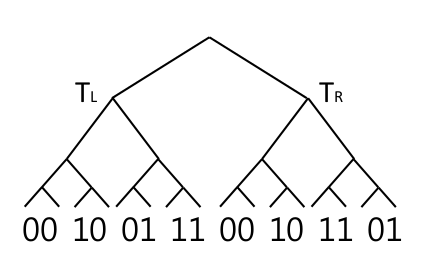}
\caption{Binary string represented by
binary tree.}
\label{fig:binary tree}
\end{figure}

\begin{table}[tbp] \centering%
\caption{Rewriting Rules for the binary tree in Fig.2.}%
\begin{tabular}{|l|l|}
\hline
$P\rightarrow \lbrack -FT_{L}][+FT_{R}]$ & $T_{R}\rightarrow \lbrack
-FT_{R_{L}}][+FT_{R_{R}}]$ \\ 
$T_{L}\rightarrow \lbrack -FT_{L_{L}}][+FT_{L_{R}}]$ & $T_{R_{L}}\rightarrow
\lbrack -FT_{R_{L_{L}}}][+FT_{R_{L_{R}}}]$ \\ 
$T_{L_{L}}\rightarrow \lbrack -FT_{L_{L_{L}}}][+FT_{L_{L_{R}}}]$ & $%
T_{R_{L_{L}}}\rightarrow \lbrack -F][+F]$ \\ 
$T_{L_{L_{L}}}\rightarrow \lbrack -F][+F]$ & $T_{R_{L_{R}}}\rightarrow
\lbrack -F][+F]$ \\ 
$T_{L_{L_{R}}}\rightarrow \lbrack -F][+F]$ & $T_{R_{R}}\rightarrow \lbrack
-FT_{R_{R_{L}}}][+FT_{R_{R_{R}}}]$ \\ 
$T_{L_{R}}\rightarrow \lbrack -FT_{L_{R_{L}}}][+FT_{L_{R_{R}}}]$ & $%
T_{R_{R_{L}}}\rightarrow \lbrack -F][+F]$ \\ 
$T_{L_{R_{L}}}\rightarrow \lbrack -F][+F]$ & $T_{R_{R_{R}}}\rightarrow
\lbrack -F][+F]$ \\ 
$T_{L_{R_{R}}}\rightarrow \lbrack -F][+F]$ &  \\ \hline
\end{tabular}

\label{tab:rewriting rules}%
\end{table}%

\subsection{Classifying rewriting rules into different sets}

Two similarity definitions are used in classifying the rewriting rules.

\begin{definition}
Homomorphism in rewriting rules. We say rewriting rule $R_1$ and rewriting
rule $R_2$ are homomorphic to each other if and only if they have the same
structure.
\end{definition}

\begin{definition}
Isomorphism on level X in rewriting rules. Rewriting rule $R_1$ and
rewriting rule $R_2$ are isomorphic on depth $X$ if they are homomophic and
their non-terminals are relatively isomophic on depth $X-1$. Isomorphic on
level 0 indicates homomorphism.
\end{definition}

After defining the similarity between rules by homomorphism and isomorphism,
we can classify all the rules into different subsets where each subset has
the same similarity relation. We will use the rule name as the class name.
For example, we asign the terminal rewriting rule a class, \textquotedblleft 
$C_{3}\rightarrow $ null\textquotedblright . Assign a rule linked to two
terminals, \textquotedblleft $C_{2}\rightarrow C_{3}C_{3}$\textquotedblright
, here $C_{3}$ is the terminal class. After classification, we obtain a
context free grammar set, which can be converted into an automata. After
transforming the binary tree in Fig,~\ref{fig:binary tree} into the set of
rewriting rules in Table~\ref{tab:rewriting rules}, we can do classification
and get the results listed in Table~\ref{tab:parameter}.

\begin{table}[tbp] \centering%
\caption{Classification based on the similarity of rewriting rules.
Several parameters for nodes and subtrees are attached to their symbols.}%
\begin{tabular}{llll}
\hline
& Classification of Rules & Isomorphic Depth \#2 & Substring \\ \hline
$\left( n=10\right) $ & Class \#1 $\left( n_{1}=3\right) $ & $n_{11}$ \ \ \
\ \ $\ a_{111}a_{112}$ &  \\ 
&  & ( 1)$C_{1}\rightarrow C_{1}C_{1}$ &  \\ 
&  & $n_{12}$ \ \ \ \ \ $\ a_{121}a_{122}$ & 00100111 \\ 
&  & ( 1)$C_{1}\rightarrow C_{4}C_{3}$ &  \\ 
&  & $n_{13}$ \ \ \ \ \ $\ a_{131}a_{132}$ & 00101101 \\ 
&  & ( 1)$C_{1}\rightarrow C_{4}C_{2}$ &  \\ \hline
& Class \#2 $\left( n_{2}=1\right) $ & $n_{21}$ \ \ \ \ \ $\ a_{211}a_{212}$
& 1101 \\ 
&  & ( 1)$C_{2}\rightarrow C_{5}C_{7}$ &  \\ 
& Class \#3 $\left( n_{3}=1\right) $ & $n_{31}$ \ \ \ \ \ $\ a_{311}a_{312}$
& 0111 \\ 
&  & ( 1)$C_{3}\rightarrow C_{7}C_{5}$ &  \\ 
& Class \#4 $\left( n_{4}=1\right) $ & $n_{41}$ \ \ \ \ \ $\ a_{411}a_{412}$
& 0010 \\ 
&  & ( 2)$C_{4}\rightarrow C_{8}C_{6}$ &  \\ 
& Class \#5 $\left( n_{5}=1\right) $ & $n_{51}$ \ \ \ \ \ $\ a_{511}a_{512}$
& 11 \\ 
&  & ( 2)$C_{5}\rightarrow C_{9}C_{9}$ &  \\ 
& Class \#6 $\left( n_{6}=1\right) $ & $n_{61}$ \ \ \ \ \ $\ a_{611}a_{612}$
& 10 \\ 
&  & ( 2)$C_{6}\rightarrow C_{9}C_{10}$ &  \\ 
& Class \#7 $\left( n_{7}=1\right) $ & $n_{71}$ \ \ \ \ \ $\ a_{711}a_{712}$
& 01 \\ 
&  & ( 2)$C_{7}\rightarrow C_{10}C_{9}$ &  \\ 
& Class \#8 $\left( n_{8}=1\right) $ & $n_{81}$ \ \ \ \ \ $\ a_{811}a_{812}$
& 00 \\ 
&  & ( 2)$C_{8}\rightarrow C_{10}C_{10}$ &  \\ 
& Class \#9 $\left( n_{9}=1\right) $ & $n_{91}$ \ \ \ \ \ $\ a_{911}a_{912}$
& 1 \\ 
&  & ( 8)$C_{9}\rightarrow $ null &  \\ 
& Class \#10 $\left( n_{10}=1\right) $ & $n_{10,1}$ \ \ \ \ $%
a_{10,11}a_{10,12}$ & 0 \\ 
&  & ( 8)$C_{10}\rightarrow $ null &  \\ \hline
\end{tabular}%
\label{tab:parameter}%
\end{table}%


\subsection{Complexity for classified rules}

The generating function of a context free grammar is defined in the
following paragraph.

\begin{definition}
Generating function of a context free grammar.

\begin{enumerate}
\item Assume that there are $n$ classes of rules, $\{C_{1},C_{2},\dots
,C_{n}\}$, and the class $C_{i}$ contains $n_{i}$ rules. Let $V_{i}\in
\{C_{1},C_{2},\dots ,C_{n}\}$, $U_{ij}\in \{R_{ij},i=1,2,\dots ,n;$ $%
j=1,2,\dots ,n_{i}\}$, and $a_{ijk}\in \{x:x=1,2,\dots ,n\}$, where each $%
U_{ij}$ has the following form for a binary tree: 
\[
\begin{array}{rl}
U_{i1} & \rightarrow V_{a_{i11}}V_{a_{i12}} \\ 
U_{i2} & \rightarrow V_{a_{i21}}V_{a_{i22}} \\ 
\dots & \rightarrow \dots \\ 
U_{in_{i}} & \rightarrow V_{a_{i{n_{i}}1}}V_{a_{i{n_{i}}2}}\text{.}%
\end{array}%
\]

\item The generating function of $V_{i},V_{i}(z)$ has a form, 
\[
V_{i}(z)=\frac{\sum%
\limits_{p=1}^{n_{i}}n_{ip}z^{k}V_{a_{ip1}}(z)V_{a_{ip2}}(z)}{%
\sum\limits_{q=1}^{n_{i}}n_{iq}}\text{.} 
\]%
If $V_{i}$ does not have any non-terminal, we set $V_{i}(z)=1$. The
parameter $k$ is set to $1$, $k=1$, in \cite{Liou10}. An alternative setting
is to weight the redundancy of rewriting rules by setting $k=1/n_{ip}$.


\item After formulating the generating function $V_{i}(z)$, we plan to find
the largest value of $z,z^{max}$, where $V_{1}(z^{max})$ is convergent. Note
that we will use $V_{1}$ to denote the function of the root node of the
binary tree. After obtaining the largest value, $z^{max}$, of $V_{1}(z)$, we
set $R=z^{max}$, where $R$ is the radius of convergence of $V_{1}(z)$. The
complexity, $K_{0}$, of the binary tree is 
\[
K_{0}=-\ln R\text{.} 
\]

\item Since computing the maximum value, $z^{max}$, directly is not
feasible, we use iterations and region tests to accomplish the complexity.
Rewrite the generating function in an iterative form, 
\[
\begin{array}{rl}
V_{i}^{m}(z^{\prime }) & =\frac{\sum\limits_{p=1}^{n_{i}}n_{ip}z^{\prime
k}V_{a_{ip1}}^{m-1}(z^{\prime })V_{a_{ip2}}^{m-1}(z^{\prime })}{%
\sum\limits_{q=1}^{n_{i}}n_{iq}}\quad ;\text{ }m=1,2,3,...;\text{ }\mbox{and}
\\ 
V_{i}^{0}(z^{\prime }) & =1\text{.}%
\end{array}%
\]

\item The value of the function $V_{i}$ at a specific $z^{\prime }$ can be
calculated by iterations of the form. In each iteration, calculate the
values from $V_{i}^{0}(z^{\prime })$ to $V_{i}^{m}(z^{\prime })$. When $%
V_{i}^{m-1}(z^{\prime })=V_{i}^{m}(z^{\prime })$ is satisfied for all rules,
we \ stop the iteration. From experiences, we set $m=200$.

\item Now we can test whether $V_{i}(z^{\prime })$ is convergent or
divergent at a value $z^{\prime }$. We use binary seaching to test the
values between $0$ and $1$. In each test, when $V_{i}(z^{\prime })$ is
convergent, we set a bigger value $z^{\prime }$ in the next test. When $%
V_{i}(z^{\prime })$ is divergent, we set a smaller value $z^{\prime }$ in
the next test. We expect that this test will approach the radius, $R=z^{max}$%
, closely.
\end{enumerate}
\end{definition}


\section{Complexity of encoded text}

We show how to compute the complexity of the text. A text sequence is first
transformed into a binary string by any giving encoding method. One can
directly set each character be an integer and obtain a binary string for the
text. For example, use the integer indexes, $1$ to $27$, to represent the 26
alphabets plus the space character. We use the term BIN to call this
encoding method. This method is simple and doesn't apply any sophisticated
encoding algorithm. A different method, Lempel-Ziv-Welch (LZW), is also
applied to encode the text. LZW is designed for lossless data compression
and is a dictionary-based encoding \cite{Welch84}. In LZW, when certain
substring appears frequently in the text, it will be saved in the
dictionary. These two methods will be used in this work to accomplish the
binary strings.

Before LZW processing, its dictionary contains all possible single
characters of the text. LZW searches through the text sequence,
successively, for a longer substring until it finds one that is not in the
dictionary. Whenever LZW finds a substring that is in the dictionary, its
index is retrieved from the dictionary and this index will replace the
substring's place in the encoded sequence. LZW will add a new substring to
the dictionary and attach a new index to the substring. The last character
of the newly replaced substring will be used as the next starting character
to scan for new substrings. Longer strings are saved, successively, in the
dictionary and made available for subsequent encoding. When LZW works on
sequence with many repeated patterns, its compression efficiency is high.

For example, suppose there are only three characters \textquotedblleft
a\textquotedblright , \textquotedblleft b\textquotedblright ,
\textquotedblleft c\textquotedblright\ in the dictionary before we start
searching. The indeices, \textquotedblleft 1\textquotedblright ,
\textquotedblleft 2\textquotedblright , and \textquotedblleft
3\textquotedblright , are used to represent them respectively. Giving a text
\textquotedblleft abcabcabc\textquotedblright , the substrings,
\textquotedblleft ab\textquotedblright , \textquotedblleft
bc\textquotedblright , \textquotedblleft ca\textquotedblright ,
\textquotedblleft abc\textquotedblright , \textquotedblleft
cab\textquotedblright , will be saved in the dictionary successively with
their new assigned indeices, \textquotedblleft 4\textquotedblright ,
\textquotedblleft 5\textquotedblright , \textquotedblleft
6\textquotedblright , \textquotedblleft 7\textquotedblright ,
\textquotedblleft 8\textquotedblright . This string \textquotedblleft
abcabcabc\textquotedblright\ will be transformed into an array
[1,2,3,4,6,5]. By using binary numbers, the array [1,2,3,4,6,5] can be
transformed into a binary string "001 010 011 100 110 101".

%
%
%

\bigskip

\noindent%
\large \textbf{Example}%

\bigskip

The article \textquotedblleft The Declaration of
Independence\textquotedblright\ is used as the text sequence. After removing
all punctuation, the total number of characters are 7930. Apply the two
encoding methods and obtain its two binary strings. We calculate the
complexity every 512 bits along the string. Figure 3 shows that the
complexity values of BIN are roughly fixed across the text sequence. The
complexity values of LZW near the front end of the text are lower than those
near the rear end of the text. Those lower valuses reveals the encoding
features of LZW. Since the LZW dictionary saves a lot of regular patterns in
the front end and absorbs the regularity, there will be no such regular
patterns in the rear end. A string with high regularity has low complexity.
So, the represented string near the rear end becomes much random with high
complexity.



%
%


\begin{figure}[tbp]
\centering
\includegraphics[width=4.0157in,height=2.846in]{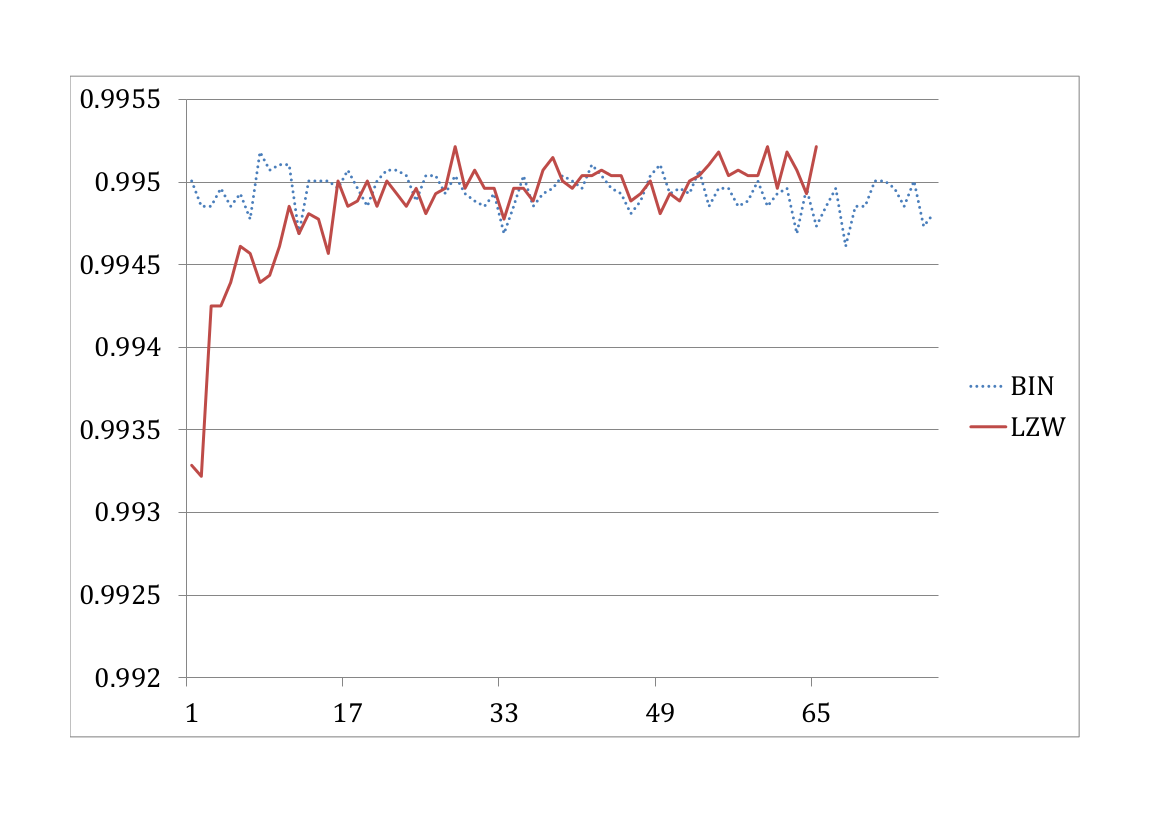}
\caption{The complexity of the
Declaration of Independence}
\label{}
\end{figure}

%
%
%
%
%
%

\section{Comparison with other Measures}

Two other measures of complexity, topological entropy (TE) \cite{Koslicki11}
and linguistic complexity (LC) are discussed and compared with the proposed
method.

\bigskip

\noindent%
\large \textbf{Topological entropy}%

\bigskip

An information function $A_{i}(s)$ is defined as, 
\[
A_{l}(s)=|\{u:|u|=l\text{ and }u\text{ is a distinct substring in }s\}|\text{%
,} 
\]%
where $A_{i}(s)$ represents the total number of distinct substrings with
length $l$ in the sequence $s$. The entropy \cite{Kirillova00} is defined
as, 
\[
H_{l}(s)=\frac{log_{k}A_{l}(s)}{l}\text{,} 
\]%
where $k$ is the size of all alphabets of the sequence. Since there are many
values for $l$ and this definition can't produce a single value of
complexity for the whole sequence. A new definition \cite{Koslicki11} for
the topological entropy is in the following paragraph.

\begin{definition}
Let $s$ be a finite sequence of length $|s|$ and $k$ be the size of
alphabet, let $l$ be the unique integer such that

\[
k^{l}+l-1\leq |s|\leq k^{l+1}+(l+1)-1 
\]%
We use $s_{1}^{k^{l}+l-1}$ to represent the first $k^{l}+l-1$ letters of s.


\[
H_{TE}(s):=\frac{log_{k}(A_{l}(s_{1}^{k^{l}+l-1}))}{l} 
\]%
where $A_{l}(s_{1}^{k^{l}+l-1})$ is the number of distinct substrings with
length $l$ in sequence $s_{1}^{k^{l}+l-1}$.
\end{definition}

\bigskip

\noindent%
\large \textbf{Linguistic complexity}%

\bigskip

Linguistic complexity \cite{GabrielianB99,Trifonov90} is a measure of the
vocabulary richness of a text. LC is defined as the ratio of the number of
substrings presented in the string of interest to the maximum number of
substrings of the same length string over the same alphabet. Thus, the more
complex a text sequence, the richer its vocabulary, whereas a repetitious
sequence has relatively lower complexity. Some notations are used in LC. Let 
$|s|$ be the length of the binary string $s$. Let $M_{l}(s)$ denote the
maximal possible number of distinct substrings with length $l$, and $%
M_{l}(s) $ is equal to $min(2^{i},|s|-i+1)$. Let $M(s)$ be the sum of $%
M_{1}(s)$, $M_{2}(s)$,.., and $M_{|s|}(s)$. Let $A_{l}(s)$ be the actual
number of distinct substrings with length $l$. Let $A(s)$ be the sum of $%
A_{1}(s)$, $A_{2}(s)$,..., and $A_{|s|}(s)$. Then, LC of the sequence $s$ is 
$A(s)/M(s)$.

For example, consider the binary string \textquotedblleft $s_{1}:0111001100$%
\textquotedblright . The length of $s_{1}$ is equal to $|s_{1}|=10$. The $%
M(s_{1})$\ value is $M(s_{1})=2+4+8+7+6+5+4+3+2+1=42$. Note that this value
depends only on the length of the string and the alphabet size. We then use
the suffix tree to calculate the actual number of distinct substrings, $%
A(s_{1})=38$. The LC value of $s_{1}$ is equal to $38/42=0.904762$.

We take another example \textquotedblleft $s_{2}:0101010101$%
\textquotedblright . The length of $s_{2}$ is equal to $10$, and the value $%
M(s_{2})=42$ is also equal to $M(s_{1})$. The $A(s_{2})$\ value $A(s_{2})=19$
is smaller than that of $A(s_{1})$. This is because $s_{2}$ is more regular.
The LC of $s_{2}$ is equal to $19/42=0.45238$.

\bigskip

\noindent%
\large \textbf{Comparison and analysis}%

\bigskip

Topological entropy focuses on only one subword length and computes its
complexity. In contrast, linguistic complexity computes the complexity of
all possible subword lengths. LC uses much more computations than that of
TE. The proposed method uses the binary tree to represent a binary sequence
that has a length of power of $2$. It reveals the structural information of
the sequence.

\begin{figure}[tbp]
\centering
\includegraphics[width=4.0157in,height=2.846in]{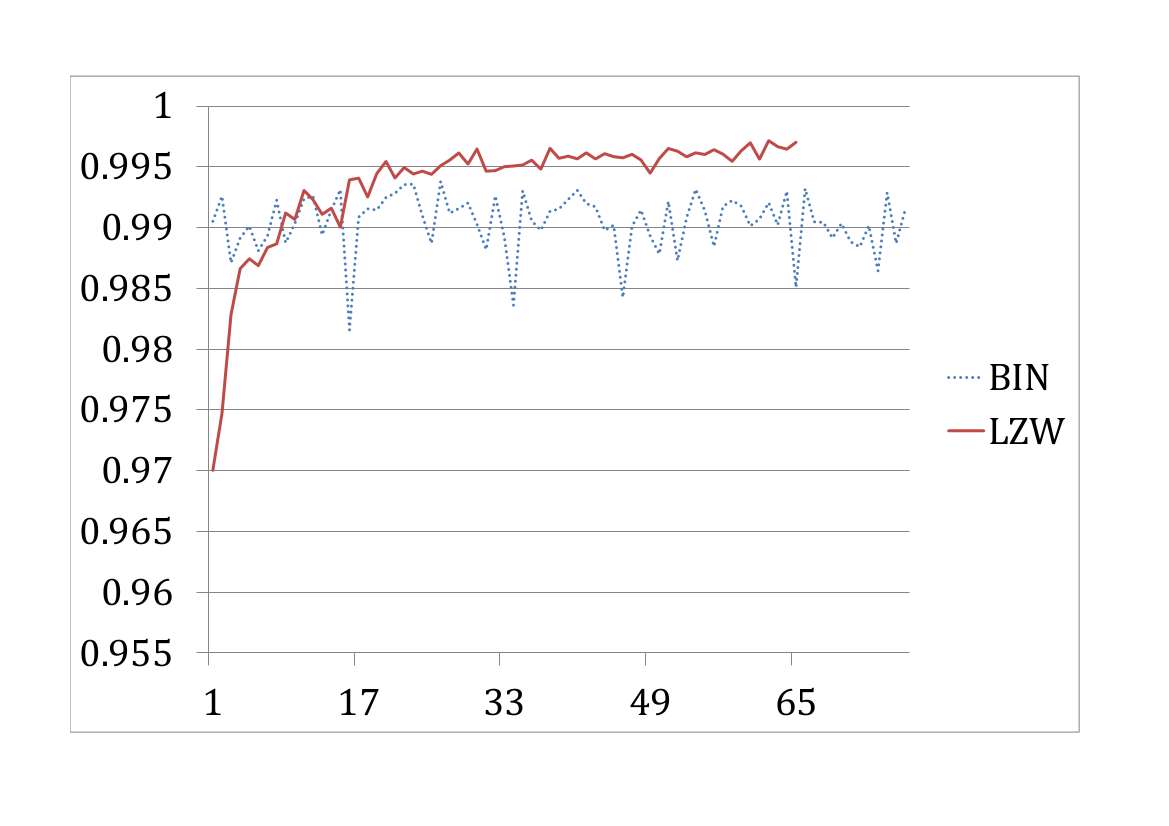}
\caption{The linguistic complexity of the
Declaration of Independence}
\label{}
\end{figure}

\begin{figure}[tbp]
\centering
\includegraphics[width=4.0157in,height=2.846in]{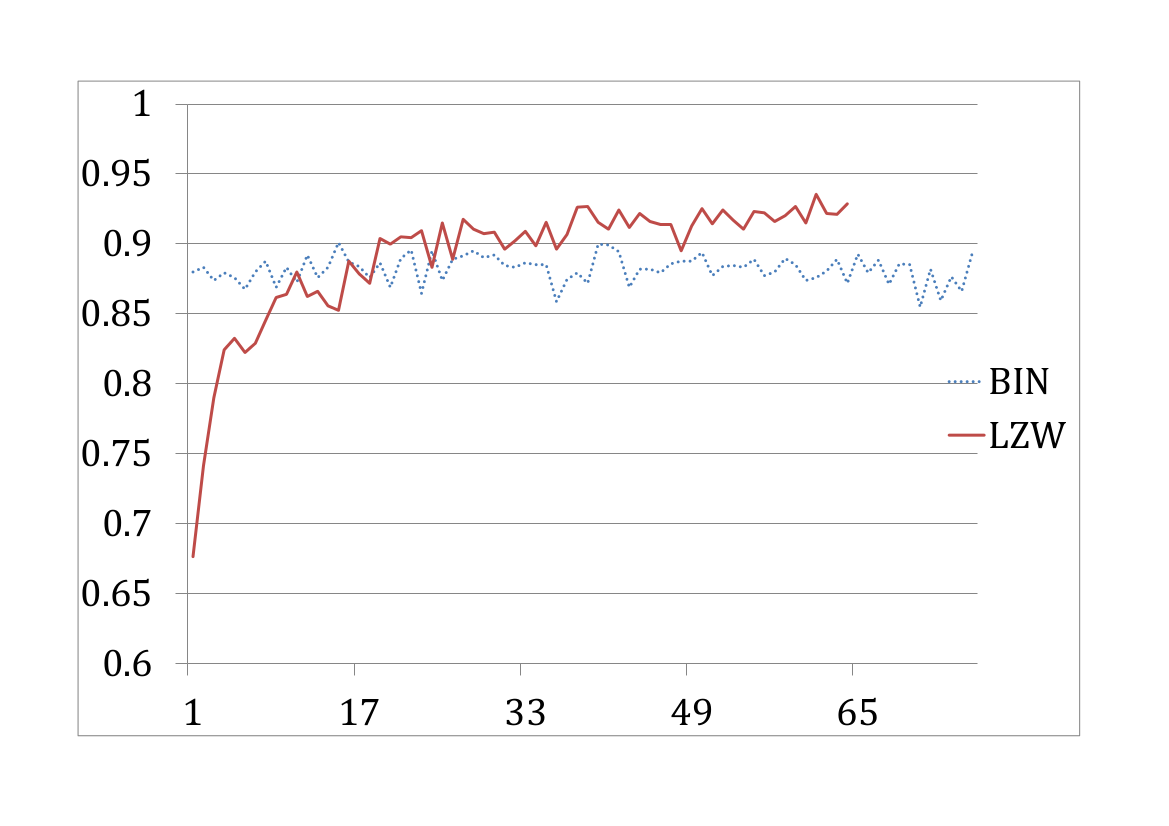}
\caption{The topological entropy of the
Declaration of Independence}
\label{}
\end{figure}


\label{c:conc}

Finally, we discuss some potential applications. The proposed complexity can
be used to assist encoding and data compression. By monitoring the
complexity of a text sequence, we can encode certain text sections of low
complexity with better compression ratios and with fewer bits. We can also
use the proposed complexity to detect anomaly in data transmission.

\bibliographystyle{splncs}
\bibliography{bibfile}

\end{document}